# Residual CNDS

*Hussein A. Al-Barazanchi, Hussam Qassim, and Abhishek Verma*
Department of Computer Science
California State University
Fullerton, CA 92834, U.S.A.
hussein@nidabasys.com, hualkassam@csu.fullerton.edu, averma@fullerton.edu

*Abstract— Convolutional Neural Networks nowadays are of tremendous importance for any image classification system. One of the most investigated methods to increase the accuracy of CNN is by increasing the depth of CNN. Increasing the depth by stacking more layers also increases the difficulty of training besides making it computationally expensive. Some research found that adding auxiliary forks after intermediate layers increases the accuracy. Specifying which intermediate layer should have the fork just addressed recently. Where a simple rule were used to detect the position of intermediate layers that needs the auxiliary supervision fork. This technique known as convolutional neural networks with deep supervision (CNDS). This technique enhanced the accuracy of classification over the straight forward CNN used on the MIT places dataset and ImageNet. In the other side, Residual Learning is another technique emerged recently to ease the training of very deep CNN. Residual learning framework changed the learning of layers from unreferenced functions to learning residual function with regard to the layer's input. Residual Learning achieved state of arts results on ImageNet 2015 and COCO competitions. In this paper, we study the effect of adding residual Connections to CNDS network. Our experiments results show increasing of accuracy over using CNDS only.*

## I. INTRODUCTION

ILSVRC contest [1] is the current test bed for computer vision algorithms. Convolutional neural networks have achieved breakthroughs in this competition [2] and also in other image classification tasks [3, 4]. Convolutional neural networks layers learn the images' low, mid, and high level features and classifiers [5] in an end to end framework. The quality of these features' levels can be boosted by the number of used layers in the network. In the ILSVRC contest, it was revealed that the convolutional neural network's accuracy can get better by increasing the network depth (number of layers) [6, 7]. This shows that the depth of the network is of critical importance. Foremost results obtained by [6, 7, 8, 9] all use very deep convolutional neural networks models on the ImageNet dataset [10]. The benefits of very deep models can be extended from regular image classification tasks to other significant recognition challenges such as object detection and segmentation [11, 12, 13, 14, 15]. On the other hand, increasing the depth of the network by adding more layers to the network will also increase the number of parameters which makes the convergence of back-propagation very slow and also prone to overfitting. Also, increasing the depth will make the gradients vulnerable to the issue of vanishing/exploding of gradients [16, 17].

Using of the pre-trained weights of shallower networks to initialize the weight of deeper networks was proposed by Simonyan and Zisserman [6]. The proposed solution is to solve the problem of slower convergence and overfitting. Nevertheless, using this technique to train multiple networks of incremental depth is computationally expensive and may result difficulty in tuning the parameters. Another technique to overcome this challenge proposed by Szegedy et al. [7]. Where they used subsidiary branches attached to the middle layers. These subsidiary branches are auxiliary classifiers. The goal of the Szegedy et al [7] of using these classifiers is to increase the gradients propagate back through layers of the deep neural network structure. Also to use these branches to motivate the feature maps in the shallower layers to anticipate the labels used in the final layer. Still, they did not specify a method that can determine the location of where to add these branches or how to add them. Lee et al. [18] follow similar idea by proposing to add the subsidiary branches after each intermediate layer. The losses from these branches are summed with the loss of the final layer. This technique showed an enhancement in the rate of convergence. They did not explore the deeply supervised networks [18] with very deep networks.

Wang et al. [19] suggested convolutional neural networks with deep supervision. Where they addressed the issue of where to add the auxiliary branches. They used the method of exploring vanishing gradients in deep networks to determine which intermediate layer needs to have an auxiliary branch. Adding auxiliary addresses the problem of slower convergence and overfitting. Even though the network is now able to start converging another challenge surface which is degradation problem. The degradation problem still expose with deeper networks especially when increasing the depth. Degradation issue start saturating the accuracy of the network and force it to break down quickly. Surprisingly, overfitting is not the reason behind the degradation problem. Degradation lead to a higher training error as reported by [20, 21] when extending the network depth by adding more layers. Moreover, the degradation that happen to the accuracy during the training phase shows that not every neural networks are equivalently easy to optimize. Residual learning [22] is a recently developed technique that solve the issue of degradation. In our paper, we handle issues of overfitting and degradation together by combining the CNDS network with Residual Learning. We add residual connections [22] into the basic CNDS [19] 8 layers structure. Our design shows that the resulted network benefit from both structures and enhance the accuracy over CNDS network.

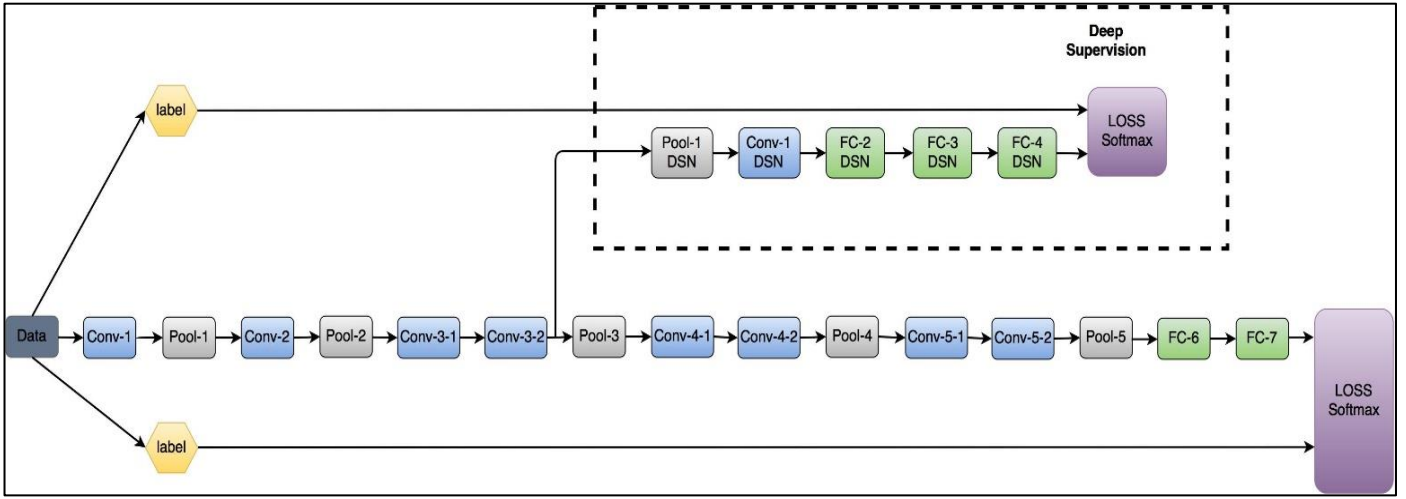

Figure 1. The CNDS Network [19].

The rest of this paper is organized as follows. In section II, we give a brief background about the CNDS network and residual learning. We discuss the details of our method Residual CNDS in section III. In section IV, we present the used data set for our experiment. The discussion of the results are in section V. We conclude the paper and suggest future work in section VI.

## II. BACKGROUND

The idea of very deep neural network was began in the 2014 ILSVRC contest [1]. As a result, the research on how to train such networks is recently started to be solved. In this section we will explore the architecture of CNDS network and how the authors used vanishing gradients to locate the best position to add the auxiliary branch. Also, the residual learning technique is explained in B. It is essential to understand the structure of both networks and how it helped us to merge them.

### A. CNDS Network

Adding auxiliary classifiers that connect to some layers in the middle was introduced by Szegedy et al. [7]. These extra classifiers helps the generalization of the network by allowing extra supervision in the training phase. Anyway, Szegedy et al. [7] procedure does not provide any rules for where the auxiliary classifiers should be connected and what is their depth. In the other side, Lee et al. [18] provides more thoroughly analysis of where to insert auxiliary classifiers branches. In their network deeply-supervised nets (DSN), they insert SVM classifier on the output of every hidden layer. They use this technique only during the training phase. Their optimization is on the sum of the output layer's loss besides the losses of auxiliary classifiers.

Wang et al. [19] addressed the issue of where to insert the auxiliary branches in their network convolutional neural networks with deep supervision (CNDS). CNDS network [19] has some major differences from the Lee et al. [18] network. The first difference is Lee et al. [18] insert auxiliary classifier after every hidden layer. In the other side, Wang et al. [19] use gradient-based heuristic rule to decide where they should connected the supervision branches. The rule is described below. The second difference is Wang et al. [19] model uses a small neural networks as an auxiliary classifiers. This small network contains convolutional layer, some fully connected layers, and end up with a softmax layer where it is very similar to [7]. Whilst Lee et al. [16] used SVM classifiers that are connected to the outputs of each hidden layer.

Wang et al. [19] built their decision of where to put the auxiliary supervision classifiers (branches) based on the vanishing of gradients. First, they designed the network without any branch. The weights are initialized from Gaussian distribution with zero mean, std=0.01, and bias=0. After that, they execute several backpropagation (10-50) epochs and they monitored the average gradient values of middle layers by plotting them. Then they decided to add the auxiliary classifier branch where the average gradient value degrades (get lower than $10^{-7}$ (threshold)). In their design, in the fourth convolutional layer the mean gradient get lower than the specified threshold. As a result, they added the auxiliary branch after the fourth layer. Figure-1 shows Wang et al. [19] resulted design. In Figure-1, we changed the naming conventions of convolutional layers to follow the naming guidelines of [6]. So the fourth layer is named conv3_2 in Figure 1.

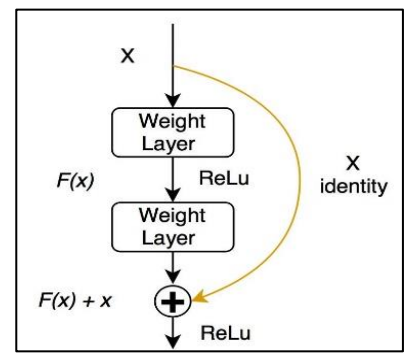

Figure 2. Residual Connection [22].

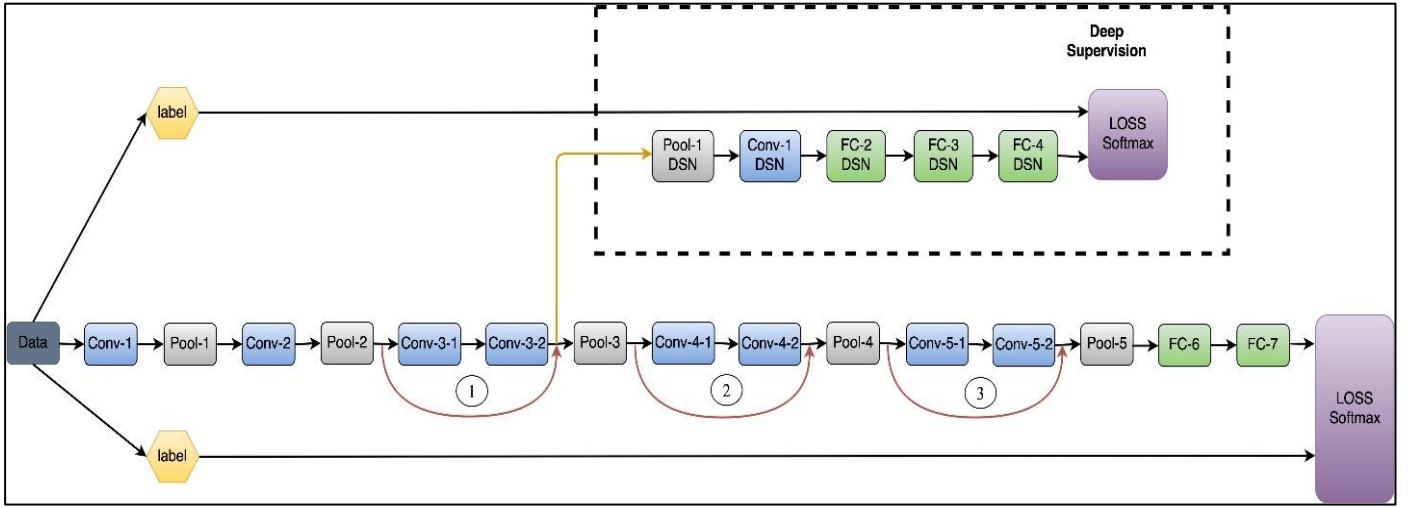

Figure 3. CNDS Network with Residual Connections (Residual CNDS).

## B. Residual Learning

Not all convolutional neural networks are easy to optimize because of the degradation problem. Increasing the depth of the network is supposed to increase the resulted accuracy. But the experimental results show different case where deeper networks produce an error higher than their counterpart shallower networks. He et al. [22], solved the issue of degradation by proposing deep residual learning. He et al. [22], let each few stacked layers fit residual mapping rather than waiting for these layers to fit the wanted mapping. Where fitting this wanted underlying mapping is not happening because of the degradation. Where they changed the underlying fitting mapping to be as in equation 2 instead of being as in equation 1. They hypothesize that it is more difficult to optimize the original mapping than to optimize the residual mapping.

$$F(x) = H(x) \quad (1)$$

$$F(x) = H(x) - x \ [22] \quad (2)$$

$$F(x) = H(x) + x \ [22] \quad (3)$$

In feedforward neural networks, shortcut connections can be used to represent equation 3. Shortcut connections is the operation of skipping one or more layers in the network [23, 24, 25]. Figure 2 shows how the shortcut connection can be implemented in CNN. As shown in Figure 2, He et al. [22] used the shortcut connections to implement identity mapping [22]. The output that is resulted from these shortcut connections is summed with the output resulted from the stacked layers as illustrated in Figure 2 which represent equation 3. The benefits of [22] identity shortcut connections is they are parameters free and they add only negligible amount of computation for the summation operation. This is in contrast with "highway networks" [21] where they introduced shortcut connections with gating functions [26] where these gates have parameters. The other benefit of identity shortcut connections of [22] is they can be optimized by SGD algorithm in an end to end manner. Also, they are easily to implement using available deep learning libraries such as [27, 28, 29, 30].

## II. RESIDUAL CNDS

The CNDS network is composed from 8 convolutional layers in the main branch. The kernel size of the first layer is 7*7 with a stride of 2. While the kernel size for the rest of layers is 3*3 with a stride of 1. Wang et al. [19] rule is add the auxiliary supervision classifier is after the layer that is suffering from gradients vanishing. So in this network architecture they found that the auxiliary supervision branch should be inserted after layer four. Feature maps generated from lower convolutional layers are noisy and it is necessary to reduce this noise before feeding it to the classifiers. To reduce the noise, Wang et al. [19] reduced the dimensionality of the features maps and passed them in non-linear functions before inserting them in the classifiers. Based on this, the auxiliary branch begin with pooling layer (Average pooling layer) of size 5 with a stride of 2. After that, there is convolutional layer of kernel size 1 and stride 1. Then they inserted two fully connected layers each of them of size 1024 and followed by 0.5 dropout ratio. While the main branch contains also two fully connected layer but of size 4096 which also followed by 0.5 dropout ratio. Each branch, the main branch and the auxiliary branch, is connected by an output layer which is connected by a softmax layer to compute the probability of classes. Figure 1 shows the CNDS network with minor changes to layers naming convention.

$$W\_main = (W1, ..., W11) \ [19] \quad (4)$$

$$W\_branch = (Ws5, ..., Ws8) \ [19] \quad (5)$$

CNDS network as mentioned before is composed from main and auxiliary branch as shown in Figure 1. Let's assume the naming of weights in main branch is as shown in equation 4. These weights correspond to the 8 convolutional layers and 3 fully connected layers. On the other hand, the auxiliary branch weights are as shown in equation 5 where the weights correspond to 1 convolutional layer and 3 fully connected layers. If we consider the feature map produced from the output layer in main branch is X11 then computing the probability using the softmax function for the labels k =1, ..., K is as in equation 6. In the other side, considering the feature produced from the output layer in the auxiliary branch is S8, then computing the response is as shown in equation 7.

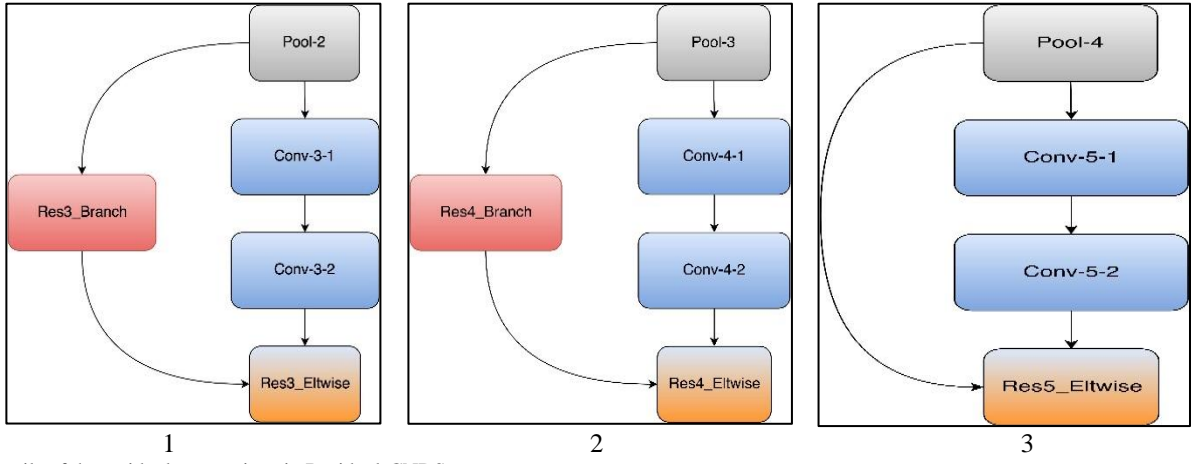

Figure 4. Details of the residual connections in Residual-CNDS.

$$pk = \frac{exp(X_{11(k)})}{\sum_k exp(X_{11(k)})} \ [19] \qquad (6)$$

$$psk = \frac{exp(S_{8(k)})}{\sum_k exp(S_{8(k)})} \ [19] \qquad (7)$$

Computing the loss for the main branch using the probabilities generated from the softmax function is shown in equation 8. While computing the loss for the auxiliary branch is shown in equation 9. Computing the loss for the auxiliary branch includes the weights from the auxiliary branch and the weights from early convolutional layers in the main branch.

$$L_0(W\_main) = - \sum_{k=1}^{K} yk \ln pk \ [19] \qquad (8)$$

$$L_s(W\_main, W\_branch) = - \sum_{k=1}^{K} yk \ln psk \ [19] \qquad (9)$$

The loss from the two branches, main and auxiliary, is combined using the formula shown in equation 10. This equation is a weighted sum where the main branch is given more weight than the auxiliary branch. The term $α_t$ is used to control the importance of the auxiliary branch as a regulizer. The usage of the $α_t$ follow [18] to be decayed with epochs as shown in equation 11 where N is the total number of epochs.

$$L_s(W\_main, W\_branch) = L_0(W\_main) + α_t L_s(W\_main, W\_branch) \ [19] \quad (10)$$

$$α_t = α_t * (1 - t/N) \ [19] \qquad (11)$$

The shortcut connections from residual learning [22] are used in our model as shown in equation 12.

$$y = F(x, \{W_i\}) + x \ [22] \qquad (12)$$

After analyzing the CNDS network, we decided to add the residual learning connections [23] only in the main branch. The places that we can add the residual connections to are the positions that have consecutive convolutional layers without pooling layers in between. So for this reason we cannot add residual connections to the auxiliary branch because it has only one convolutional layer. Figure 3 shows the resulted architecture after adding residual connections to CNDS network. As shown in Figure 3, the first residual connection connects the input to the Conv3-1 with the output from Conv3-2. Where the element-wise addition connects the output of Pool-2 with output of the convolutional layer (Conv3-2). The number of kernels in Conv-2 is 128 while the number of kernels in Conv3-2 is 256. To make the number of kernels' outputs equal so we can perform the element-wise addition, we added a convolution layer of 256 kernels between Pool-2 and the element-wise addition. This is shown in Figure 4-1. Res3_Branch represents the convolutional layer that is added.

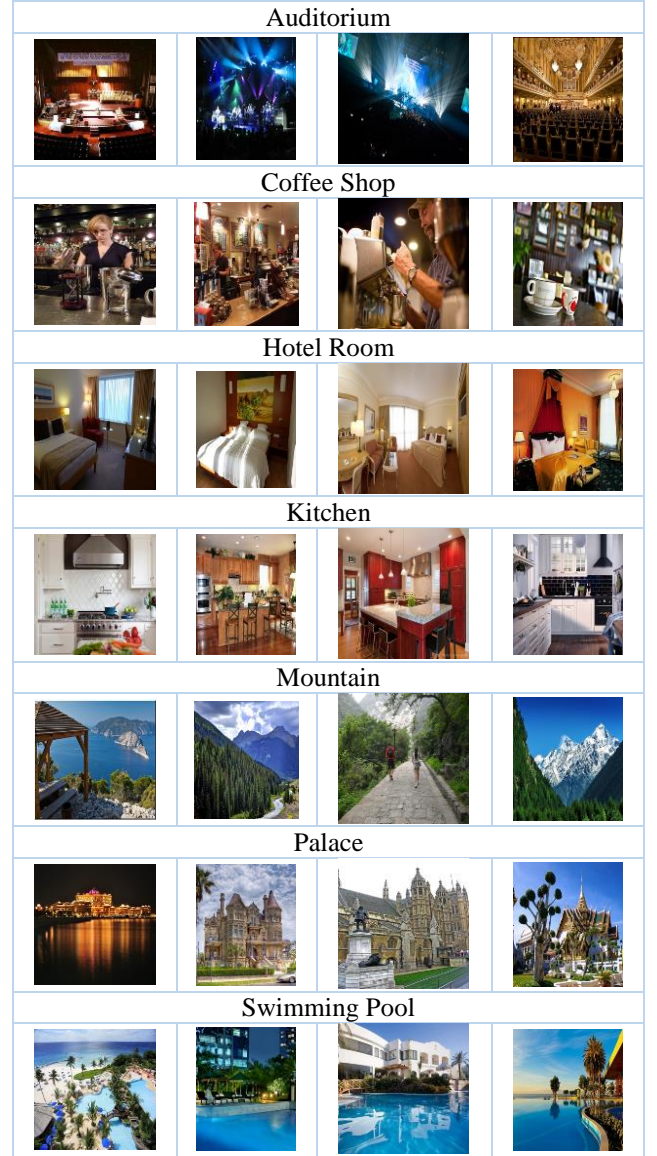

Figure 5. Random samples from MIT Places.

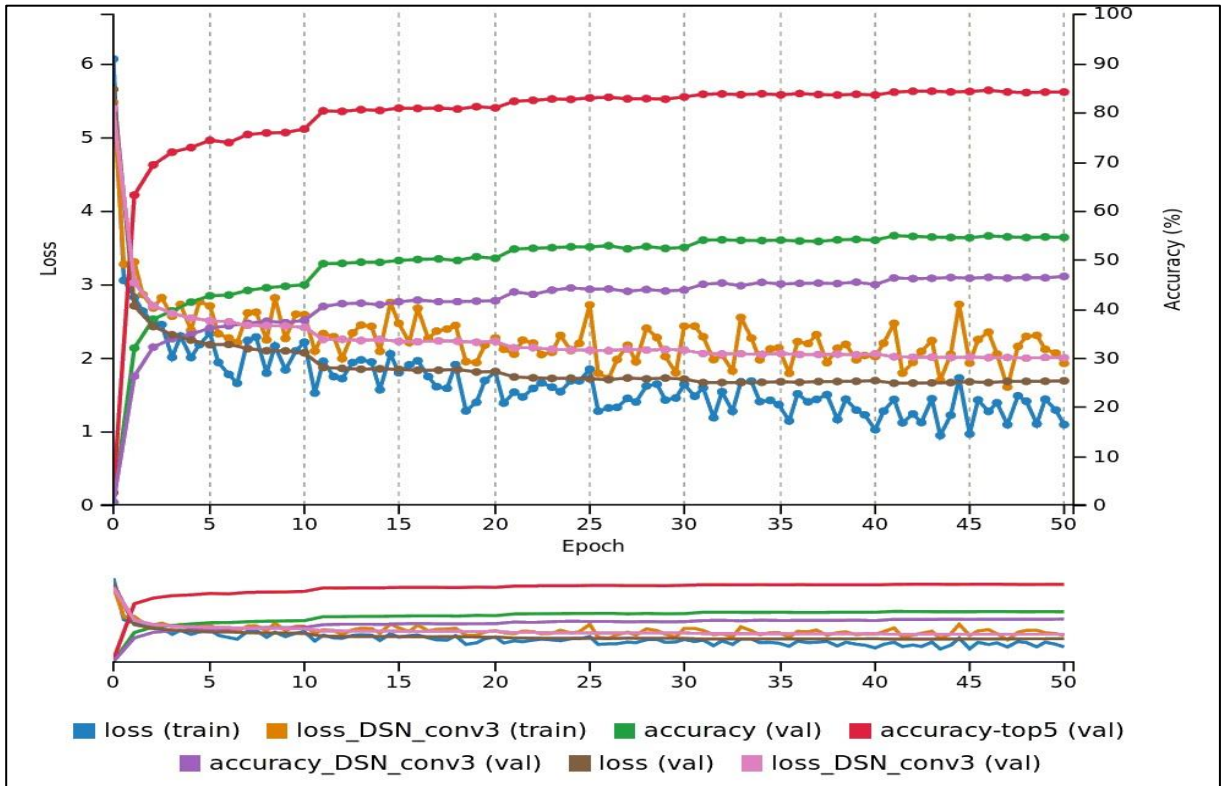

Figure 6. Loss and Accuracy

The second residual connection added is after Pool-3 layer and the shortcut connection passes two convolutional layers. Thus, the residual connection is between the output of Pool-3 and the output of convolutional layer Conv4-2. We also added convolutional layer of 512 kernel between Pool-3 and before the element-wise addition to make the number of kernels of Pool-3 and Conv4-2 equal because the number of kernels of Conv3-2 is 256 while the number of kernels of Conv4-2 is 512. Figure 4-2 shows this addition. The auxiliary branch is inserted after the merging operation between the output of Pool3 and Conv4-2. Finally, the third and the last residual link connects between the output of Pool-4 and the last convolutional layer (Conv5-2). We did not need to add a convolutional layer between the Pool-4 and the element-wise addition because the number of kernels of Conv4-2 and Conv5-2 is equal, which is 512 kernels for both of them.

## IV. IMAGE DATASET

We conducted our experiments on MIT Places [31] dataset because in CNDS network they reported their result using the MIT Places [31] dataset. In this way, we can compare our results using Residual-CNDS with the CNDS network. MIT Places [31] is a large dataset created by MIT Computer Science and Artificial Intelligence Laboratory, which is bigger than ILSVRC [1] and the SUN dataset [32]. The MIT places is also called Places-205 because it contains 205 scene categories. Places-205 is scene-centric database, which is provided to help the academic experiment and education objective in the area of computer vision. This dataset contains 2.4M training images. The number of images for the 205 class is between 5000-15000 images per category. The validation set contains 100 images per category and this means that the total number of images for the validation set is 20500 image. The test set contains 200 images per category so the total number is 41000 image for testing. Figure 5 shows 4 random samples from MIT places for few classes.

## V. EXPERIMENT SETUP

In this work, we trained our model, which contains 8 convolutional layers in the main branch with the residual connections and 1 convolutional layer in the auxiliary branch, from the scratch using Caffe [28] on NVIDIA DIGITS platform [33] (Deep Learning GPU Training System). NVIDIA DIGITS [33] is an open source project that let the user built and design their neural networks jobs in the field of image classification and object detection quickly with real-time visualization. The hardware configuration of the system is 4 GPU NVIDIA GeForce GTX TITAN X. The system have 2 Xeon processors and 256 GB or memory. All images in the training, validation, and testing are resized to 256*256. The only pre-processing we implemented is subtracting the mean pixel for each color channel (RGB). We set the batch size for training to be 256 and for validation to be 128. We set the number of epochs to be 50 and the initial learning rate is 0.01. The learning rate is decreased 5 times during the training process after each 10 epochs. We set the decreasing of the learning rate to be to the half of the previous value. Images are cropped to 227*227 from random points before feeding them to the first convolutional layers. All layers weights initialized from Gaussian distribution with 0.01 standard deviation. The only image augmentation that we used is image reflection for the training data.

Table I. Classification Accuracy

| Method | Top-1 Val/Test | Top-5 Val/Test |
|---|---|---|
| AlexNet | - /50.0 | - / 81.1 |
| GoogLeNet | - / 55.5 | - / 85.7 |
| CNDS | 54.7 / 55.7 | 84.1 / 85.8 |
| Residual CNDS | 54.8/56.3 | 84.5/ 86.0 |

We trained our Residual-CNDS network on the Places-205 [31] dataset with 2.4M training images for the 205 scene classes, with 5000-15000 images per class. We validated our model with 100 images per class. The training process using the aforementioned system took two days and 14 hours. One of the powerful features of NVIDIA DIGITS is the real time visualization of the network in the training phase. Figure 6 shows the loss and the accuracy of our model. We report top-1 and top-5 accuracy on the validation and test dataset. We selected the epoch with highest accuracy to be used with the test set. In our case epoch 46 was the highest. Also, the auxiliary branch is removed during the testing and it is only used during the training. In testing, we used the average of 10-crops as used in [2]. For the test dataset, the classes' label are not available. Alternatively, we submitted our predictions to the MIT Places server to obtain the results.

## VI. DISCUSSION

In this paper, we have worked on the concept of merging two powerful ideas which are convolutional neural network with deep supervision [19] and deep residual learning [22] to train deep neural network. Adding residual connections to the CNDS network was to see if these shortcut connections can enhance the performance of the original network. Adding the residual connections to the network did not increase the complexity of the model except for a negligible amount. Experimenting on the same dataset used in testing CNDS network have supported our hypothesis that adding the residual connections to the CNDS will enhance the performance of the network.

Referring to the table I, we can see out that our model (Residual_CNDS) passed the top-1 and top-5 accuracy of the original CNDS [19] network. Our model outperform the original CNDS [18] by 0.1% in top-1 validation, 0.6% test. Furthermore we surpasses the CNDS [18] model in top-5 by 0.4% validation, and 0.2% on the test. Our model also exceed the two models (GoogLeNet [7] and AlexNet [2]) that was trained by the MIT team [31]. Table II, shows the accuracy for the top 50 class using Residual-CNDS.

## VII. CONCLUSION AND FUTURE WORK

Residual learning is a very powerful technique to train very deep neural networks. The most interesting about residual connections is they increase the complexity of the model very little. We used these shortcut connections with an already tested model called CNDS. Our experiment shows that the residual connections enhanced the performance of the CNDS network. This concludes that residual connections are useful for other networks. For future work, we recommend to study the effect of residual connections on other popular networks such as the VGG network.

Table II. Top 50 Classes

| Class | % | Class | % | Class | % | Class | % | Class | % |
|---|---|---|---|---|---|---|---|---|---|
| outdoor | 96 | pulpit | 88 | nursery | 83 | alley | 79 | outdoor | 76 |
| runway | 95 | raft | 88 | igloo | 82 | rock_arch | 79 | amphitheater | 75 |
| cockpit | 94 | phone_booth | 87 | coral_reef | 82 | aquarium | 78 | corridor | 75 |
| wind_farm | 94 | ballroom | 86 | bookstore | 81 | closet | 78 | gas_station | 75 |
| bus_interior | 92 | music_studio | 86 | bowling_alley | 81 | crosswalk | 78 | auditorium | 73 |
| fire_escape | 92 | playground | 86 | campsite | 81 | dam | 78 | cemetery | 73 |
| iceberg | 91 | shower | 86 | laundromat | 81 | water_tower | 78 | engine_room | 73 |
| lighthouse | 91 | bamboo_forest | 85 | rice_paddy | 81 | airport_terminal | 77 | sea_cliff | 73 |
| football | 91 | racecourse | 85 | fire_station | 80 | hotel_room | 77 | shoe_shop | 73 |
| boxing_ring | 89 | badlands | 83 | pagoda | 80 | martial_arts_gym | 77 | supermarket | 73 |

# REFERENCES

[1] J. Deng, A. Berg, S. Satheesh, H. Su, A. Khosla and L. Fei-Fei, "Imagenet large scale visual recognition competition 2012 (ilsvrc2012)", 2012.

[2] A. Krizhevsky, I. Sutskever, and G. E. Hinton. "Imagenet classification with deep convolutional neural networks". In Advances in neural information processing systems, pages 1097–1105, 2012.

[3] Y. LeCun, B. Boser, J. S. Denker, D. Henderson, R. E. Howard, W. Hubbard, and L. D. Jackel. "Backpropagation applied to handwritten zip code recognition". Neural computation, 1989.

[4] P. Sermanet, D. Eigen, X. Zhang, M. Mathieu, R. Fergus, and Y. LeCun. "Overfeat: Integrated recognition, localization and detection using convolutional networks". In ICLR, 2014.

[5] M. D. Zeiler and R. Fergus. "Visualizing and understanding convolutional neural networks". In ECCV, 2014.

[6] K. Simonyan and A. Zisserman. "Very deep convolutional networks for large-scale image recognition". In ICLR, 2015.

[7] C. Szegedy, W. Liu, Y. Jia, P. Sermanet, S. Reed, D. Anguelov, D. Erhan, V. Vanhoucke, and A. Rabinovich. "Going deeper with convolutions". In CVPR, 2015.

[8] K. He, X. Zhang, S. Ren, and J. Sun. "Delving deep into rectifiers: Surpassing human-level performance on imagenet classification". In ICCV, 2015.

[9] S. Ioffe and C. Szegedy. "Batch normalization: Accelerating deep network training by reducing internal covariate shift". In ICML, 2015.

[10] O. Russakovsky, J. Deng, H. Su, J. Krause, S. Satheesh, S. Ma, Z. Huang, A. Karpathy, A. Khosla, M. Bernstein, et al. "Imagenet large scale visual recognition challenge". arXiv:1409.0575, 2014.

[11] R. Girshick, J. Donahue, T. Darrell, and J. Malik. "Rich feature hierarchies for accurate object detection and semantic segmentation". In CVPR, 2014.

[12] K. He, X. Zhang, S. Ren, and J. Sun. "Spatial pyramid pooling in deep convolutional networks for visual recognition". In ECCV, 2014.

[13] R. Girshick. Fast R-CNN. In ICCV, 2015.

[14] S. Ren, K. He, R. Girshick, and J. Sun. "Faster R-CNN: Towards real-time object detection with region proposal networks". In NIPS, 2015.

[15] J. Long, E. Shelhamer, and T. Darrell. "Fully convolutional networks for semantic segmentation". In CVPR, 2015.

[16] Y. Bengio, P. Simard, and P. Frasconi. "Learning long-term dependencies with gradient descent is difficult". IEEE Transactions on Neural Networks, 5(2):157–166, 1994.

[17] X. Glorot and Y. Bengio. "Understanding the difficulty of training deep feedforward neural networks". In AISTATS, 2010.

[18] C.-Y. Lee, S. Xie, P. Gallagher, Z. Zhang, and Z. Tu. "Deeply supervised nets". arXiv preprint arXiv:1409.5185, 2014.

[19] L. Wang, C. Lee, Z. Tu, and S. Lazebnik. "Training deeper convolutional networks with deep supervision". CoRR, abs/1505.02496, 2015.

[20] K. He and J. Sun. "Convolutional neural networks at constrained time cost". In CVPR, 2015.

[21] R. K. Srivastava, K. Greff, and J. Schmidhuber. "Highway networks". arXiv:1505.00387, 2015.

[22] K. He, X. Zhang, S. Ren, and J. Sun. "Deep residual learning for image recognition". arXiv:1512.03385, 2015.

[23] C. M. Bishop. "Neural networks for pattern recognition". Oxford university press, 1995.

[24] B. D. Ripley. "Pattern recognition and neural networks". Cambridge university press, 1996.

[25] W. Venables and B. Ripley. "Modern applied statistics with s-plus". 1999.

[26] S. Hochreiter and J. Schmidhuber. "Long short-term memory". Neural computation, 9(8):1735–1780, 1997.

[27] M. Abadi, A. Agarwal, P. Barham, E. Brevdo, Z. Chen, C. Citro, G. Corrado, A. Davis, J. Dean, M. Devin, and et al. "Tensorflow: Large-scale machine learning on heterogeneous distributed systems". arXiv preprint arXiv:1603.04467, 2016.

[28] Y. Jia, E. Shelhamer, J. Donahue, S. Karayev, J. Long, R. Girshick, S. Guadarrama, and T. Darrell. "Caffe: Convolutional architecture for fast feature embedding". arXiv:1408.5093, 2014.

[29] R. Collobert, K. Kavukcuoglu, and C. Farabet. "Torch7: A MATLAB-like environment for machine learning". In BigLearn, NIPS Workshop, 2011.

[30] F. Chollet. "Keras". GitHub repository, https://github.com/fchollet/keras, 2015.

[31] B. Zhou, A. Lapedriza, J. Xiao, A. Torralba, and A. Oliva. "Learning Deep Features for Scene Recognition using Places Database". NIPS, 2014.

[32] J. Xiao, J. Hays, K. Ehinger, A. Oliva, and A. Torralba. "SUN Database: Large-scale Scene Recognition from Abbey to Zoo". IEEE Conference on Computer Vision and Pattern Recognition, 2010.

[33] NVIDIA DIGITS Software. (2015). Retrieved April 23, 2016, from https://developer.nvidia.com/digits